# Optimizing Contrail Detection: A Deep Learning Approach with EfficientNet-b4 Encoding

Qunwei Lin*
Trine University
Phoenix, USA
linqunwei1030@outlook.com

Qian Leng
University of Maryland Eastern Shore
Princess Anne, USA
ql150@georgetown.edu

Zhicheng Ding
Columbia University
New York, USA
zhicheng.ding@columbia.edu

Chao Yan
Northeastern University
Boston, USA
ycmike97@gmail.com

Xiaonan Xu
Northern Arizona University
Flagstaff, USA
xiaonan.xu.academic@outlook.com

*Abstract*—In the pursuit of environmental sustainability, the aviation industry faces the challenge of minimizing its ecological footprint. Among the key solutions is contrail avoidance, targeting the linear ice-crystal clouds produced by aircraft exhaust. These contrails exacerbate global warming by trapping atmospheric heat, necessitating precise segmentation and comprehensive analysis of contrail images to gauge their environmental impact. However, this segmentation task is complex due to the varying appearances of contrails under different atmospheric conditions and potential misalignment issues in predictive modeling. This paper presents an innovative deep learning approach utilizing the efficientnet-b4 encoder for feature extraction, seamlessly integrating misalignment correction, soft labeling, and pseudo-labeling techniques to enhance the accuracy and efficiency of contrail detection in satellite imagery. The proposed methodology aims to redefine contrail image analysis and contribute to the objectives of sustainable aviation by providing a robust framework for precise contrail detection and analysis in satellite imagery, thus aiding in the mitigation of aviation's environmental impact.

## I. INTRODUCTION

The global imperative for environmental sustainability has propelled industries, particularly the aviation sector, to reassess their operational strategies to mitigate their ecological impact. Aviation, a cornerstone of modern global connectivity, is increasingly scrutinized for its significant contribution to environmental challenges, particularly concerning greenhouse gas emissions and their role in climate change. Among the multifaceted challenges facing the aviation industry, contrails, the linear ice-crystal clouds trailing behind aircraft, have garnered attention due to their significant impact on global warming. As aircraft engines emit exhaust gases, including water vapor and particles, they create contrails that trap heat in the Earth's atmosphere, exacerbating the greenhouse effect and contributing to climate change.

Contrails, initially perceived as benign phenomena, have come under scrutiny for their environmental ramifications, necessitating a comprehensive understanding of their formation, persistence, and overall impact on the climate. Precise segmentation and analysis of contrail images are imperative to assess their environmental implications accurately. However, this task is beset with complexities stemming from the dynamic nature of contrails under varied atmospheric conditions, potential misalignment issues in predictive modeling, and the inherent variability in human annotations.

Addressing these challenges requires innovative approaches that leverage advanced technologies and methodologies. In this context, deep learning emerges as a powerful tool for contrail image analysis, offering the potential to redefine the landscape of contrail detection and analysis in satellite imagery. By harnessing the capabilities of deep learning architectures, such as the efficientnet-b4 encoder, and integrating cutting-edge techniques like misalignment correction, soft labeling, and pseudo-labeling, it becomes possible to enhance the accuracy, efficiency, and robustness of contrail detection algorithms.

This paper presents a groundbreaking deep learning approach tailored explicitly for contrail detection in satellite imagery, aimed at addressing the complexities inherent in contrail segmentation tasks. The proposed methodology not only seeks to refine contrail image analysis but also contributes to the broader objectives of sustainable aviation practices by providing a framework for accurate contrail detection and analysis. By advancing the state-of-the-art in contrail detection, this research endeavors to support efforts towards mitigating aviation's environmental impact and fostering a more sustainable future for air travel.

## II. RELATED WORK

The landscape of contrail image segmentation has been shaped by an extensive body of research that spans foundational approaches and recent advancements.

Krizhevsky et al. [1] were pioneers in demonstrating the potency of deep convolutional neural networks (CNNs) for image classification, laying the foundation for subsequent advancements in segmentation methodologies. A comprehensive survey by Zaitoun et al. [2] charted the trajectory of image segmentation techniques, tracing its journey from classical methods right up to modern deep learning paradigms. Grady et al. [3] offered an alternative lens to view image segmentation, introducing a probabilistic perspective by formulating segmentation as a random walk on a graph. Zhao et al. [4] introduces a self-supervised active learning framework for biomedical image segmentation, leveraging SSL features for sample selection without initial labels, leading to improved performance and lower annotation costs in skin lesion segmentation. Zhang et al. [5] introduces a novel federated learning approach to enhance system robustness against Byzantine attacks while maintaining stringent privacy guarantees.

Zhang et al. [6]address ranking optimal answers to science questions from large language models, showcasing Platypus2-70B's exceptional performance with a benchmark score of 0.909904.Xu et al. [7] propose a Curriculum Recommendations paradigm utilizing innovative techniques to enhance learning equity and effectiveness, validated by competitive cross-validation scores.Xu et al. [8] explore AI's impact on tumor pathology, focusing on image and molecular marker analysis for intelligent diagnosis and treatment model development.Wang et al. [9] explore merging cloud computing and machine learning, addressing resource management and future trends.

Diving deeper into segmentation-specific methodologies, Ronneberger et al. [10] carved a niche with their U-Net architecture, originally designed for biomedical image segmentation. Fully convolutional networks (FCNs) by Long et al. [11] provided a paradigm shift, emphasizing pixel-wise semantic segmentation.Liu et al. [12] boosts erosion studies in restored plantation forests using UAV images and CNN models, improving accuracy over traditional methods. The study also offers practical control strategies. The multi-scale approach was championed by Zhao et al. [13], leading to the development of the Pyramid Scene Parsing Network. Further, Chen et al. [14] refined this direction, introducing the DeepLab architecture which adeptly captured multi-scale contextual information.

Modern research has seen an influx of novel methodologies and strategies tailored for specific challenges in contrail image segmentation. Fu et al. [15] put forth the Dual Attention Network, a blend of spatial and channel attention mechanisms. Bai et al. [16] converged deep learning with the watershed transform for nuanced instance segmentation. Emphasizing the importance of data accuracy, He et al. [17] ventured into the realm of pseudo-labeling, aligning pseudo-labels with true labels for enriched deep learning training. Recent advancements in representation learning and neural translation models have showcased their efficacy in diverse challenges. Zhao et al. [18] Study utilizes Digital Twin for urban expansion's impact on reserve vegetation, using UAV, satellite, and geographic data for sustainable development. Hassan et al. [19], on the other hand, approached contrail detection from satellite images using a dynamic programming lens. Zhang et al. [20]proposed framework utilizes dynamic convolutions and multiple object tracking philosophy, resulting in state-of-the-art performance. Chollet et al. [21] ushered in efficiency in neural network design with their introduction of depthwise separable convolutions via the Xception architecture, a pivotal development for real-time segmentation tasks.

The panorama of image segmentation research has seen a steady progression from foundational techniques to the sophistication of modern deep learning architectures. Cheng et al. [22] proposes DRNet: Two-stage framework with AODD for localization, DReID for recognition, and attribute module for accuracy. Liang et al. [23] study presents EventKITTI: A dataset for event-based traffic detection. GFA-Net and CGFA-Net, the object detectors, achieve high speed-accuracy.Xu et al. [24]propose an automated scoring system for clinical patient notes using advanced NLP techniques and pseudo labeling, aiming to improve efficiency and effectiveness in evaluation. Although architectures like Dual Attention Networks, BiSeNet, and DeepLab have been instrumental for a broad spectrum of image segmentation tasks, contrail image segmentation introduces its own set of specialized challenges.X Yan et al. [25]proposes an unsupervised deep learning method for MRI image denoising, useful for medical applications with limited paired training data.W Weimin et al. [26]presents an EAS U-Net model using deep learning to enhance liver segmentation, especially for small targets. Zhou and colleagues [27] demonstrated how pseudo-labeling could effectively utilize unlabeled datasets to refine model predictions and adjust misalignments in an unsupervised context.

Our research endeavors to address these challenges directly. By leveraging the efficientnet-b4 for feature extraction, we have tailored our approach specifically to capture the nuanced characteristics of contrail images. Additionally, our seamless amalgamation of misalignment correction, soft labeling, and the inclusion of pseudo-labels extends the boundaries of traditional methodologies. This not only helps in navigating the variability of contrail appearances but also mitigates potential predictive discrepancies.

## III. MODEL AND METHODOLOGY

### A. Feature Extraction with EfficientNet-b4

The heart of our segmentation model resides in the EfficientNet-b4 [28] architecture, which is known for its unparalleled balance between accuracy and efficiency. The driving innovations behind its prowess are:

*1) Compound Scaling:* EfficientNet, unlike traditional networks, does not scale the network's depth, width, or resolution in isolation. Instead, it introduces compound scaling—scaling all three dimensions systematically. The relationship is defined as:

$$d = \alpha^\phi \quad (1)$$
$$w = \beta^\phi \quad (2)$$
$$r = \gamma^\phi \quad (3)$$

where $d$, $w$, and $r$ denote depth, width, and resolution, respectively. The coefficient $\phi$ is user-defined, and $\alpha$, $\beta$, $\gamma$ are constants determined empirically.

*2) MBConv - Mobile Inverted Bottleneck Convolution:* Leveraging MBConv blocks, EfficientNet pushes the envelope in computational efficiency. These blocks, inspired by MobileNetV2, use inverted residuals and linear bottlenecks to reduce computational load while maintaining robustness.

*3) Squeeze-and-Excitation (SE) Block:* The SE block, integrated within EfficientNet, offers channel recalibration. By providing weighted importance to channels, SE blocks ensure that more relevant channels contribute more to the feature representation. Given a feature map $F(x)$:

$$F_{SE}(x) = \text{sigmoid}(W_2 \delta(W_1 F_{\text{avg}}(x))) \quad (4)$$

where $W_1$ and $W_2$ represent reduction and expansion operations.

*4) Swish Activation Function:* EfficientNet employs the Swish function, touted for outperforming traditional activation functions. The Swish function is:

$$\text{Swish}(x) = x \cdot \text{sigmoid}(\beta x) \quad (5)$$

where $\beta$ is learnable.

Given EfficientNet-b4's deeper architecture and enhanced specifications, it offers superior feature extraction.

### B. Misalignment Correction

Even minor misalignments in segmentation tasks can critically affect model performance. An intriguing observation during our experimentation was a perceivable drop in the cross-validation (CV) score when utilizing flip/rot90 augmentations.

*1) Root of Misalignment:* The origin of this inconsistency was hypothesized to be the conversion of polygon annotations to binary masks. The transformation seemed to inadvertently produce a misalignment between the image and its corresponding mask, with the mask excluding the leftmost point while encompassing the rightmost one.

*2) Correction Strategy:* To counteract this misalignment, a subtle +0.5 pixel shift was introduced to the image during both training and inference phases. This approach aimed to realign the image with its mask, ensuring consistent segmentation results.

Mathematically, let the predicted output be $O_p$ and the ground truth be $O_g$. The corrected output $O_c$ is derived as:

$$O_c = \mathcal{C}(O_p, O_g) \quad (6)$$

where $\mathcal{C}$ denotes the alignment correction function, implemented using an affine transformation. This correction mechanism successfully mitigated the potential degradation in model performance due to misalignment.

### C. Soft Labeling Strategy

In the realm of image segmentation, binary hard labels have been the convention for defining object boundaries. However, specific challenges, especially when nuances are paramount, necessitate a more refined labeling strategy. In our task of contrail detection, soft labels, which imbue a gradient of confidence rather than absolute certainties, have demonstrated superior efficacy in capturing the subtle variances intrinsic to contrails.

Several characteristics are pivotal when discerning contrails:

1) A spatial threshold is imposed wherein contrails must encompass no fewer than 10 pixels to be considered.
2) Geometrically, valid contrails at any given juncture in their trajectory are at least three times longer than their width.
3) In terms of appearance dynamics, contrails either manifest abruptly within an image or progress into the frame from its edges.
4) To bolster the confidence of detection, contrails should be discernible across at least two sequential images.

For the ground truth establishment, a collaborative annotation strategy was employed. Each image underwent scrutiny by more than four annotators. A pixel was deemed to be part of a contrail if over half the annotators concurred in their assessment. Interestingly, while both individual annotations and a consolidated aggregated label were made available, training using the mean of individual annotations outperformed the latter strategy. This can be attributed to the fact that averaging individual annotations yields labels that portray the level of agreement among annotators, furnishing the model with gradients of confidence. Such a paradigm avoids the pitfalls of hard binary decisions, especially in areas of contention.

Mathematically, the soft label generation is expounded by:

$$L_{soft}(p) = \frac{1}{N} \sum_{i=1}^{N} L_i(p) \quad (7)$$

where $L_{soft}(p)$ delineates the soft label of pixel $p$, $N$ signifies the total number of annotators, and $L_i(p)$ represents the annotation of the $i$-th annotator for pixel $p$.

### D. Pseudo-labels and Two-Phase Training

In the context of our segmentation methodology, Figure 1 provides an overview of our hybrid training paradigm, emphasizing transitions between the different phases. This approach capitalizes on both labeled and unlabeled data, facilitated by the use of pseudo-labels. Pseudo-labels, which are the model's own predictions on the unlabeled dataset, serve as a valuable proxy for actual labels, especially when labeled data is scarce.

**Rationale:** The underlying intuition of leveraging pseudo-labels stems from the assumption that the model, after an initial training phase, can provide reasonably accurate predictions for the unlabeled data. By using these predictions as pseudo-labels and retraining the model, we aim to amplify the confidence and accuracy of the model, especially in ambiguous regions.

**Primary Training Phase:** The first step of our two-phase approach involves traditional supervised learning. Utilizing five-fold cross-validation, the model is trained on distinct subsets of the labeled data, and its performance is evaluated on a corresponding validation set. The average error across these $k$ splits is denoted as $E_{CV}$, and is given by:

$$E_{CV} = \frac{1}{k} \sum_{i=1}^{k} E_i \quad (8)$$

Where $E_i$ represents the error for the $i$-th fold.

**Pseudo-label Generation:** After achieving convergence in the primary phase, the model is deemed ready to generate pseudo-labels. The trained model predicts labels for the unlabeled dataset $D_u$, resulting in:

$$L_p = P(D_u) \quad (9)$$

Where $P$ is the function representing the model's prediction process. **Final Training Phase:** With the pseudo-labels in hand, they are amalgamated with the original labeled dataset to form an enriched training set. The model is then retrained on this combined dataset. This not only refines the model's generalization capabilities but also ensures better consistency in its predictions, capitalizing on the insights gleaned from its own pseudo-labels.

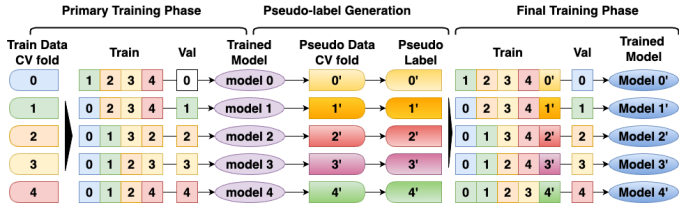

Fig. 1. Schematic of the two-phase training methodology, elucidating the transitions from primary training, pseudo-label generation, to the final training cycle.

### E. Evaluation Metric: Dice Coefficient

The success of our segmentation model heavily relies on a carefully selected evaluation metric. In our studies, we use the Dice Coefficient. This metric is a standard choice in image segmentation domain.

**Definition and Derivation:** For two sets $X$ and $Y$, the Dice coefficient $D$ defines their similarity as:

$$D(X, Y) = \frac{2 \times |X \cap Y|}{|X| + |Y|} \quad (10)$$

Where:
- $|X \cap Y|$ represents the size of the intersection of the sets.
- $|X|$ indicates the size of set $X$.
- $|Y|$ indicates the size of set $Y$.

The Dice Coefficient serves as a normalized similarity measure, with its results ranging from 0 (no overlap) to 1 (perfect overlap). In our segmentation context, the sets $X$ and $Y$ correspond to the pixels in the predicted segmentation and the ground truth, respectively.

### F. Loss Function

Our segmentation task aims at not only pixel-wise precision but also maintaining spatial coherence in the segmentation outputs. This dual aim is met using a composite loss function.

**Binary Cross Entropy (BCE):** BCE is the cornerstone for binary classification tasks. It quantifies the difference between predicted probabilities and the actual labels, ensuring accuracy at the pixel level:

$$\text{BCE}(P, GT) = -\frac{1}{N} \sum_{i=1}^{N} [GT_i \log(P_i) + (1 - GT_i) \log(1 - P_i)] \quad (11)$$

**Dice Coefficient in Loss:** Beyond achieving pixel accuracy, it's crucial for the model to generate spatially coherent segments that closely match the actual segments. To this end, we utilize the Dice coefficient in our loss function:

$$D(P, GT) = \frac{2|P \cap GT|}{|P| + |GT|} \quad (12)$$

The rationale behind using the Dice coefficient in our loss function is its sensitivity to spatial continuity in segmented outputs. A higher coefficient indicates the model's segmented prediction is in close alignment with the true segmentation, ensuring spatial coherence.

Integrating both BCE and the Dice coefficient, our loss function is:

$$L = 0.5 \times \text{BCE}(P, GT) + 0.5 \times (1 - D(P, GT)) \quad (13)$$

The weights (both set at $0.5$) ensure an equilibrium between pixel-wise accuracy and spatial coherence. Depending on the specific task's requirements, these weights can be adjusted.

### G. Results

*1) Experimental Setup:* Our experiments leveraged a meticulous setup to ensure the robustness of the results. The model's architecture, input size, loss function, and augmentations were selected after considering state-of-the-art practices in the segmentation domain. The specific parameters used are encapsulated in Table I:

TABLE I
PARAMETERS EMPLOYED IN THE EXPERIMENTAL SETUP, OPTIMIZED FOR SEGMENTATION TASKS.

| Parameter | Value |
|---|---|
| Encoder Backend | efficientnet b4 |
| Image Size | 512x512 |
| Loss Function | BCE×0.5 + Dice×0.5 |
| Augmentation | ShiftScaleRotate(p=0.6), hflip |

*2) Model Performance Measured by Dice Coefficient:* In the domain of image segmentation, a model's performance can vary significantly based on the inclusion of specific techniques. To ascertain the impact of each technique, we evaluated the model under different configurations. Table II showcases the Dice Coefficient scores for these configurations:

TABLE II
PERFORMANCE OF MODEL VARIATIONS DEMONSTRATING THE
INCREMENTAL IMPROVEMENTS WITH THE INTEGRATION OF SPECIFIC
TECHNIQUES.

| Description | Image Size | Dice Coefficient |
|---|---|---|
| Baseline | 512x512 | 0.67818 |
| Baseline + MC (Misalignment Correction) | 512x512 | 0.68131 |
| Baseline + MC + SL (Soft Label) | 512x512 | 0.69623 |
| Baseline + MC + SL + PL (Pseudo-Label) | 512x512 | 0.69735 |

The results in Table II offer compelling insights. The Misalignment Correction (MC) demonstrated a measurable improvement over the baseline. Further integration of Soft Label (SL) and Pseudo-Label (PL) techniques incrementally boosted the model's performance. This not only underscores the importance of these techniques but also paves the way for future experiments where a combination of multiple strategies can be explored for optimized performance.

## IV. CONCLUSION

In an era where the aviation industry grapples with environmental implications, the intricate issue of contrails takes center stage. It's within this context that our research breaks new ground, offering a cutting-edge deep learning methodology tailored explicitly for contrail image segmentation. Leveraging the efficientnet-b4 encoder, we have carved out a solution that's both innovative and effective. What sets our methodology apart are the meticulously integrated techniques: the misalignment correction which ensures precision in predictive analysis; soft labeling, which offers a flexible and adaptive training gradient; and pseudo-labeling, a strategy that harnesses previously unused data to refine the model's accuracy. It's a paradigm shift in how we approach and understand the environmental impact of contrails. Our research exemplifies the potential of melding innovative techniques with deep learning, underscoring a dedication to advancing both technology and sustainable aviation practices.